\documentclass[lettersize,journal]{IEEEtran}
\usepackage{amsmath,amsfonts}
\usepackage{algorithmic}
\usepackage{algorithm}
\usepackage{array}
\usepackage[caption=false,font=normalsize,labelfont=sf,textfont=sf]{subfig}
\usepackage{textcomp}
\usepackage{stfloats}
\usepackage{url}
\usepackage{verbatim}
\usepackage{graphicx}
\usepackage{cite}
\begin{document}

\title{Image Denoising via Style Disentanglement}

\author{ Jingwei Niu, Jun Cheng, and Shan Tan
\thanks{This work was supported in part by the National Natural Science Foundation of China (NNSFC), under Grant Nos. 61672253 and 62071197. \textit{(Corresponding author: Shan Tan.)}}
\thanks{Jingwei Niu, Jun Cheng and Shan Tan are with the School of Artificial Intelligence and Automation, Huazhong University of Science and Technology, Wuhan, China, (email: m202072849@hust.edu.cn, jcheng24@hust.edu.cn, shantan@hust.edu.cn).}
\thanks{Jingwei Niu and Jun Cheng are co-first authors.}
}

\markboth{Journal of \LaTeX\ Class Files,~Vol.~14, No.~8, February~2023}%
 {Shell \MakeLowercase{\textit{et al.}}: A Sample Article Using IEEEtran.cls for IEEE Journals}

\maketitle

\begin{abstract}
Image denoising is a fundamental task in low-level computer vision. While recent deep learning-based image denoising methods have achieved impressive performance, they are black-box models and the underlying denoising principle remains unclear. In this paper, we propose a novel approach to image denoising that offers both clear denoising mechanism and good performance. We view noise as a type of image style and remove it by incorporating noise-free styles derived from clean images. To achieve this, we design novel losses and network modules to extract noisy styles from noisy images and noise-free styles from clean images. The noise-free style induces low-response activations for noise features and high-response activations for content features in the feature space. This leads to the separation of clean contents from noise, effectively denoising the image. Unlike disentanglement-based image editing tasks that edit semantic-level attributes using styles, our main contribution lies in editing pixel-level attributes through global noise-free styles. We conduct extensive experiments on synthetic noise removal and real-world image denoising datasets (SIDD and DND), demonstrating the effectiveness of our method in terms of both PSNR and SSIM metrics. Moreover, we experimentally validate that our method offers good interpretability.
\end{abstract}


\section{Introduction}
\IEEEPARstart{D}{ue} to inherent physical limitations in various acquisition systems, image signals captured in real-world scenarios are often contaminated by random noise. This noise poses a significant challenge for downstream tasks such as image analysis and understanding. Therefore, image denoising plays a fundamental and critical role in low-level computer vision.

In recent years, deep learning-based methods have emerged as a promising approach for image denoising. These methods leverage powerful deep neural networks (DNNs), large-scale datasets, and advanced learning strategies to learn the mapping from degraded observations to the corresponding clean counterparts in an end-to-end manner. Compared to traditional methods, deep learning-based approaches have demonstrated superior performance. Various learning strategies have been employed, including residual learning \cite{l1}, multi-stage learning \cite{l2}, and improved loss functions \cite{l3}. In terms of network architectures, popular choices include encoder-decoder architectures \cite{l4,l5}, attention mechanisms \cite{l2,l6}, and generative adversarial networks (GANs) \cite{l3,l7}. Despite their effectiveness and power, these deep learning-based methods are often considered black-box models and the underlying denoising mechanism remains unclear, hindering the ability to understand and analyze the inner workings of these models. 

\begin{figure}[!t]
    \centering
    \includegraphics[width=\columnwidth]{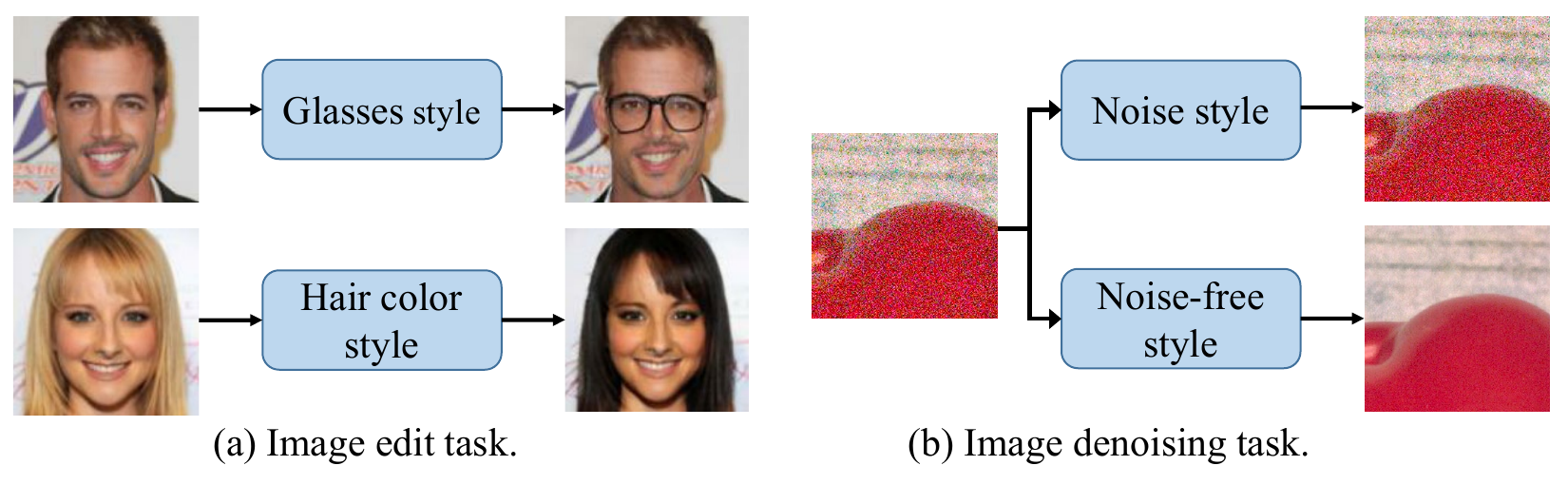}
    \caption{Style disentanglement in image editing (a) and image denoising (b). (a) In the image editing task, semantic-level attributes (e.g., the glasses style, the hair color style) are extracted and injected into the input image. These styles only affect specific parts of the image, known as local styles. (b) The image denoising task is to remove noise that is distributed across all pixels. In this case, both the noise-free style and the noise style affect pixel-level attributes.}
    \label{fig0}
\end{figure}

Disentangled representation learning (DRL), which aims to decompose variation factors in the feature representations and makes a single variation factor only affect a single image semantic attribute, is considered interpretable \cite{l8}. In this context, the semantic attribute refers to specific pixels representing a human-understandable concept, such as glasses or hair color, while variation factors are low-dimensional features corresponding to these semantic attributes in the feature space.

Currently, DRL has been applied to image processing tasks such as image editing and image reconstruction, where the cyclic consistency and style disentanglement \cite{l10,l11} are two frequently employed strategies that facilitate DRL. Cyclic consistency aims to \emph{explicitly} decouple variation factors and separate various content features in the feature space through the twice-cross-reconstruction strategy. Du et al. \cite{l10} utilized cyclic consistency for image denoising for the first time. They trained a cyclic encoder-decoder network on unpaired images and explicitly disentangled the content and noise features of noisy images and removed the noise by discarding the noise features. However, discarding imperfectly decoupled noise features may result in the loss of fine image textures and details, leading to poor image quality. On the other hand, style disentanglement decouples the latent features of images into content features and a style vector in an \emph{implicit} manner. This enables high-level image editing by injecting different style vectors into the latent features \cite{l11,l12,l13,l14}. However, image denoising requires global manipulation of pixels rather than local adjustments as in image editing. Therefore, the existing style disentanglement methods designed for general semantic editing are not directly applicable to image denoising tasks.

To this end, in this paper, we present a novel framework for image denoising, building upon the concept of style disentanglement. In our framework, we consider noise as a global image style and propose novel losses and network modules to extract noise styles from noisy images and noise-free styles from clean images. By leveraging these noise-free styles, we can adjust the latent features of noisy images, leading to effective noise removal. It is important to emphasize that our approach differs from methods used in image editing tasks, where noise-free styles are used to manipulate semantic-level attributes like glasses or hair color. Instead, we focus on pixel-level attributes, specifically noise, as illustrated in Fig. \ref{fig0}. 
In our framework, the style extractor is shared for both noisy and noise-free images, and the resulting styles are embedded in the same space. Our experiments demonstrate that the transition between noise and noise-free styles in this embedding space can generate images with varying levels of noise.  In summary, our main contributions include:
\begin{itemize}
    \item We introduce an effective and interpretable image denoising framework based on style disentanglement. 
    We consider image noise as the global image style. We design a style extractor that captures both noise and noise-free styles, and a style conversion module that utilizes noise-free styles to generate low-response activations for noise features, resulting in effective noise removal.
    \item Different from existing image denoising approaches, our method does not rely on learning complex nonlinear mappings between the degraded and clean image domains. Instead, we utilize noise-free styles to directly edit the noise features, enabling effective image denoising. 
    And our method holds the potential to provide valuable insights for safety-critical image restoration tasks.
    \item We conduct extensive experiments on synthetic noise removal and real-world image denoising to validate the effectiveness of our proposed method. Moreover, we experimentally demonstrate that our method is easier to analyze and interpret than existing end-to-end denoising networks.
\end{itemize}

\section{Related Work}
\label{gen_inst}

\subsection{Image denoising}

In recent years, DNNs-based image denoising methods have demonstrated superior performance over traditional methods \cite{l2, l3, l5, l15}.  Many network architectures and learning strategies have been developed to fit the non-linear mapping between noisy-clean image pairs. To name a few, the Unet architecture \cite{UNet}, an encoder-decoder-based approach with multi-scale feature representation capacity, is widely used in image denoising. Residual learning \cite{l1} focuses on fitting the residual signal and has been proven effective for the convergence of deeper networks. Spatial and channel attention mechanisms enable DNNs to adaptively choose key feature representations \cite{l6}. Generative adversarial strategies empower DNNs with the ability to generate visually realistic images without over-smoothing \cite{l3,l7}. Moreover, studies have shown that jointly using multiple learning strategies boosts denoising performance, such as the combination of an encoder-decoder and residual learning \cite{l5} or the incorporation of GAN and perceptual loss. Zamir et al. \cite{l2} improved image denoising performance by using multi-stage architectures, attention mechanisms, and improved losses. More recently, transformer-based image restoration models have been presented to achieve state-of-the-art performance \cite{Restormer, EEM}. However, these DNNs-based models are generally difficult to interpret, and the underlying denoising mechanism is not clear.

\begin{figure}[!t]
    \centering
    \includegraphics[width=\columnwidth]{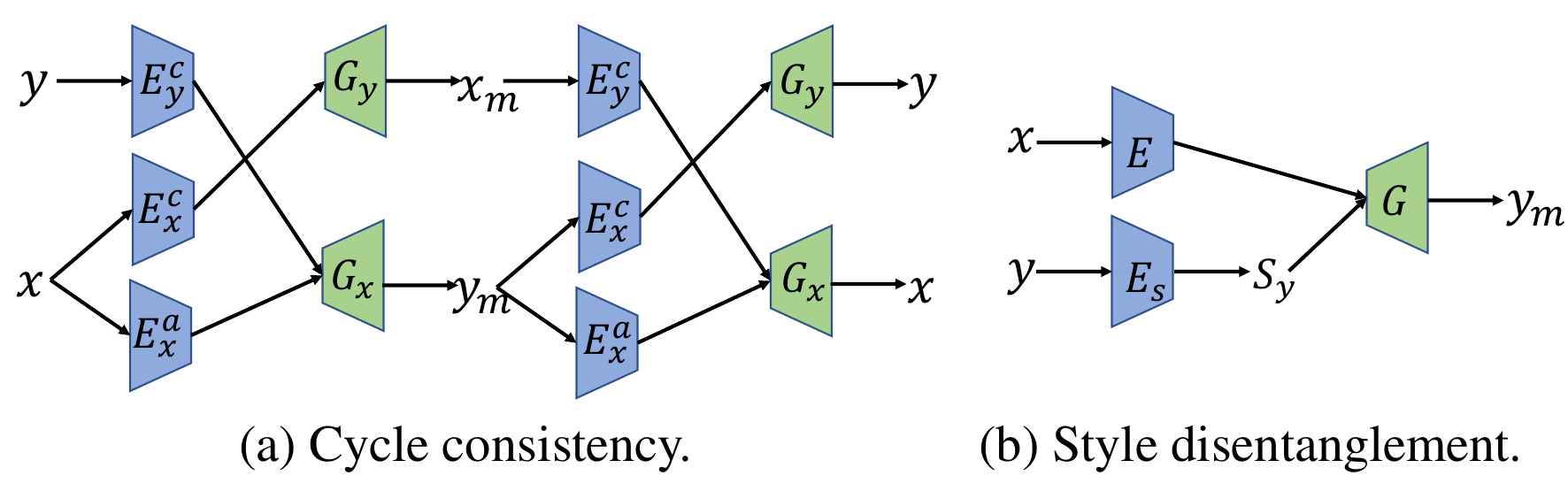}
    \caption{Diagram illustration of cycle consistency \cite{l10, l24, l25} and style disentanglement \cite{l12, l13, l14, l26, j26, j27}. (a)  Cycle consistency uses two sets of encoders to encode different features and then cross-restores the features to make the final image consistent with the input image. (b) Style disentanglement uses a style extractor to extract style vectors corresponding to semantic attributes, and then uses these style vectors to edit image features.}
    \label{fig22}
\end{figure}

\subsection{Cyclic consistency}
Cyclic consistency uses multiple encoders to encode different features of the image, and then exchanges different encoded features twice for cross-reconstruction, expecting the final reconstructed image to be consistent with the input image. Cycle consistency was first proposed within the image editing task \cite{l24}and then applied to the image restoration task \cite{l10,l25}. Lu et al. \cite{l25} utilized the cyclic consistency constraint for image deblurring for the first time. They jointly train the blur encoder $E^a$ and content encoder $E^c$ to separate the blur features and clean content features of a blurred image and then perform two cross-exchanges of the blur features to ensure consistency between the target image and the input image. Fig. \ref{fig22}(a) shows a simple diagram of cyclic consistency. Du et al. \cite{l10} further extended Lu et al.'s work to the image denoising task. However, these methods generally abandon the valuable blur or noise features, resulting in over-smoothed images.

\subsection{Style disentanglement}
Currently, style disentanglement is primarily used in image editing tasks \cite{l12, l13, l14, l26, j26, j27}, in which a specific semantic style is utilized to edit corresponding image features in a controlled manner, as shown in Fig. \ref{fig22}(b). Style disentanglement is also applied to low-level vision tasks. Zhou et al. \cite{l13} discovered that the style latent space in StyleGAN automatically encodes various semantic attributes of the image and decomposes them in a closed form. Li et al. \cite{l14} achieved the simultaneous decomposition of multiple styles in an image. Specifically, they organized multiple image tags as a hierarchical tree structure where each tag was independent, and then edited images through multiple styles.

\section{Proposed Method}
We first show the general idea of using style disentanglement for image denoising. Consider paired data sampled from the joint distribution $(x,y) \sim p_D(x,y)$, where $x$ and $y$ are from the noisy image domain and clean image domain, respectively. We assume that $x$ and $y$ share the same semantic content features $f^s$ in the latent space, while $x$ contains noise features $f^n$ that $y$ does not have, and $y$ contains clean content features $f^c$ that $x$ does not have. That is, 
\begin{equation}
\begin{split}
    (f^s,f^n) = enc(x) \\
    (f^s,f^c) = enc(y)
\end{split}
\end{equation}
where $enc$ is an encoder to extract feature representations of images. Using the corresponding decoder $dec$, features $(f^s,f^n)$ and $(f^s,f^c)$ are reconstructed into noisy images and clean images themselves:
\begin{equation}
\begin{split}
    x = dec(f^s,f^n) \\
    y = dec(f^s,f^c)
\end{split}
\end{equation}

The objective of style disentanglement is to identify domain-specific style representations. Specifically, the noise style $s_{noise}$ and the noise-free style $s_{noise\_free}$ are expected to correspond to the noisy features $f^n$ and clean content features $f^c$, respectively:
\begin{equation}
\begin{split}
    s_{noise} = ext(x)\\
    s_{noise\_free} = ext(y)
\end{split}
\end{equation}
where $ext$ is the ideal style extractor and can extract the corresponding style vector from the input image. Once the $s_{noise\_free}$ can be generated accurately, it is subsequently injected into the features of the noisy image $(f^s,f^n)$ using the style conversion module $sc$ to transform the noise features $f^n$ to $\hat{f}^c$, which approximates clean content features $f^c$:
\begin{equation}
    (f^s,\hat{f}^c) = sc((f^s,f^n), s_{noise\_free})
\end{equation}
where $s_{noise\_free}$ is passed through a fully connected layer to obtain adaptive parameters $\{s_s,s_b\}$, which are used to channel-wise modulate the features and then the style converter uses adaptive instance normalization (AdaIN) operations \cite{l11} to edit image features:
\begin{equation}
    AdaIN(e_i,s) = s_s \frac{e_i - \mu (e_i)}{\sigma (e_i)} + s_b
\end{equation}
where $e_i$ is the $i$-th channel feature, $\mu (e_i)$ and $\sigma (e_i)$ its mean and variance, respectively; $s_s$ is the scale parameter and $s_b$ is the bias parameter. 

Finally, the modified features $(f^s,\hat{f}^c)$ are used to complete the image denoising task:
\begin{equation}
    \hat{y} = dec(f^s,\hat{f}^c)
\end{equation}
where $\hat{y}$ is the expected denoised image.

Specifically, we propose a novel denoising framework called Style Disentanglement for Image Denoising Network (SDIDNet), which mainly involves noise and noise-free style extraction (SE), style generation (SG), and style conversion (SC). As shown in Fig. \ref{fig1}, an encoder is firstly used to map noisy images $x$ into a low-dimensional feature space. Meanwhile, we propose a style extractor to extract  $s_{noise}$ and $s_{noise\_free}$ from noisy and clean images under the guidance of the dedicated loss, respectively. We assume that $s_{noise}$ and $s_{noise\_free}$ can represent $f^n$ and $f^c$, respectively. Then, the SC module uses the $s_{noise\_free}$ to edit the noisy features, aiming to remove noise and preserve clean content in the feature space. Finally, a decoder is used to reconstruct the image from the edited features. The style-changed outputs of the decoder are exactly the denoised results. In this pipeline, clean images that provide noise-free styles are indispensable. Unfortunately, clean images are inaccessible in the inference phase. To address this issue, we design a style generator that takes a normal distribution as input and is trained to generate sampling styles that are similar to real noise-free styles. 

\begin{figure}[!t]
    \centering
    \includegraphics[width=0.7\columnwidth]{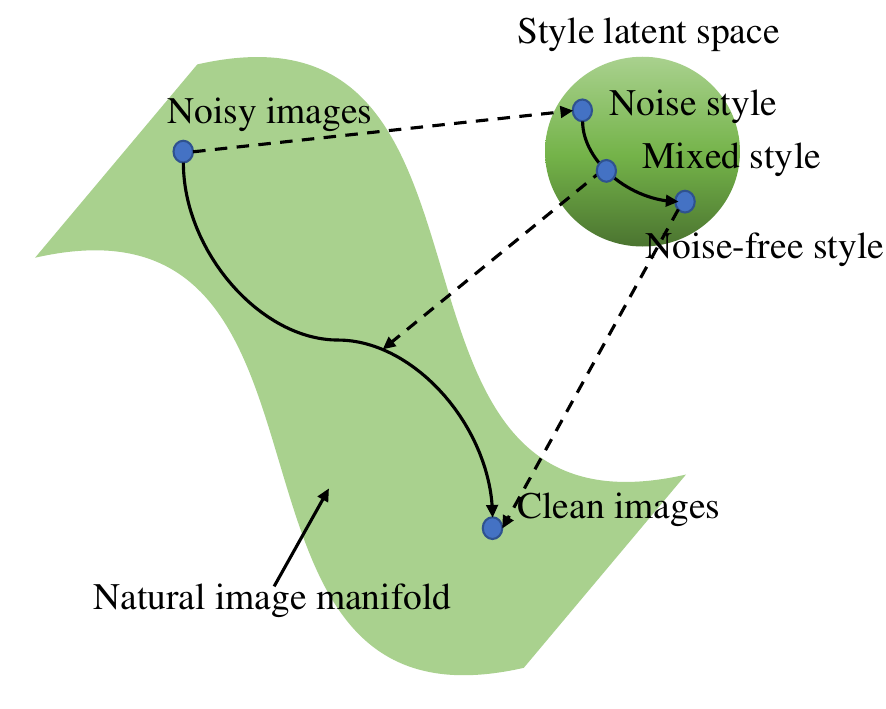}
    \caption{Schematic diagram of natural image manifold and style latent space. }
    \label{figlatent}
\end{figure}

\begin{figure*}[h]
    \centerline{\includegraphics[width=0.85\textwidth]{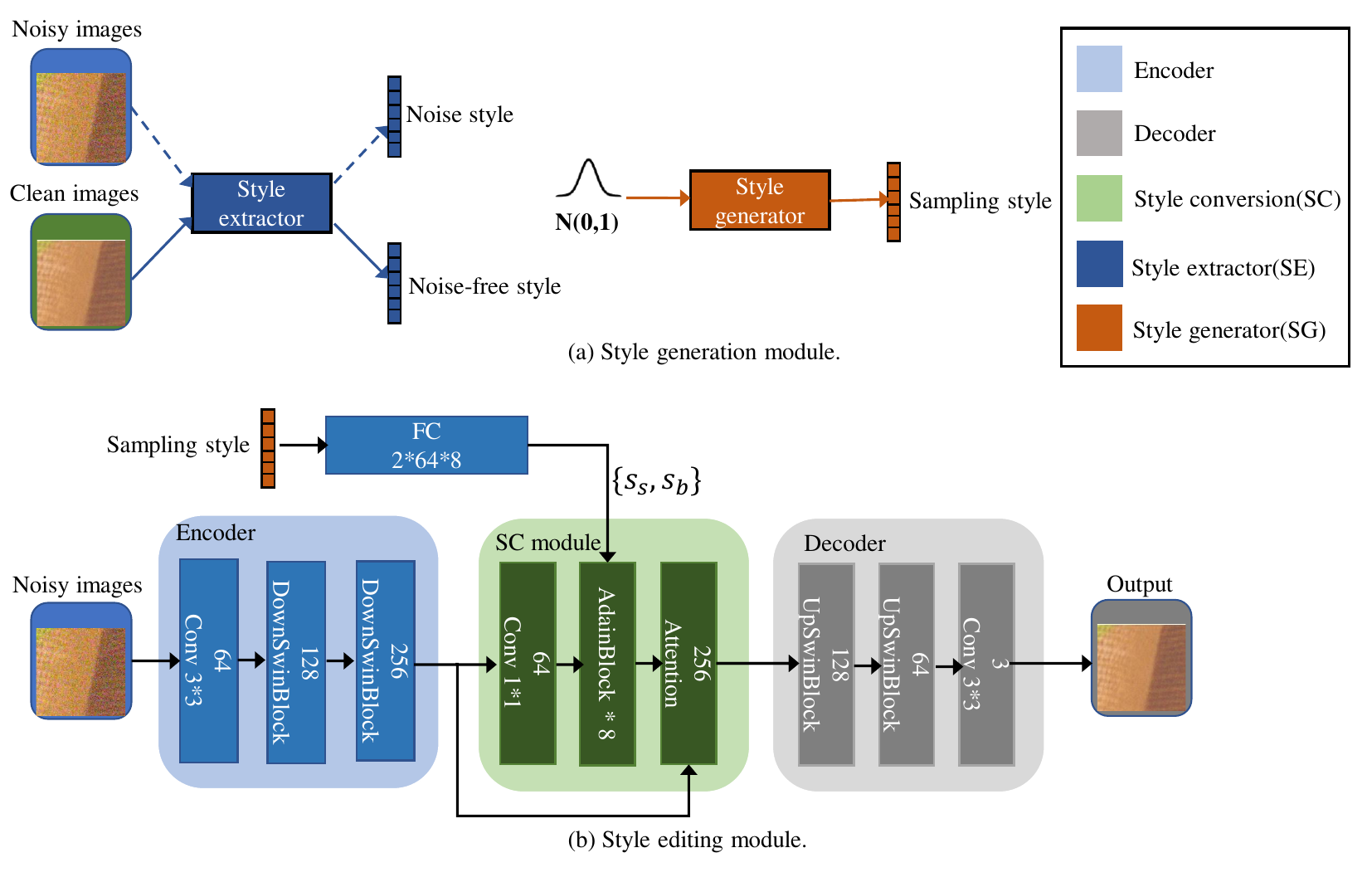}}
    \caption{The SDIDNet structure consists of (a) the style generation module and (b) the style editing module. The style generation module captures the style corresponding to the unique features, while the style editing module converts the input image's unique features into the unique features corresponding to the given style. \textbf{ Training phase:} We first encode the noisy and clean images into features using the encoder. The style extractor extracts noisy and noise-free styles from the noisy image and clean image, respectively, while the style generator generates sampling styles that mimic noise-free styles. Subsequently, the style conversion (SC) module edits noisy image features with the given styles and the noise will be removed under the guidance of the noise-free style. Finally, the decoder reconstructs the denoised image. 
    \textbf{Inference phase:} Only noisy image and sampling styles are needed to complete image denoising.}
    \label{fig1}
\end{figure*}

\label{trans}
\begin{figure}[h]
    \centering
    \includegraphics[width=0.9\columnwidth]{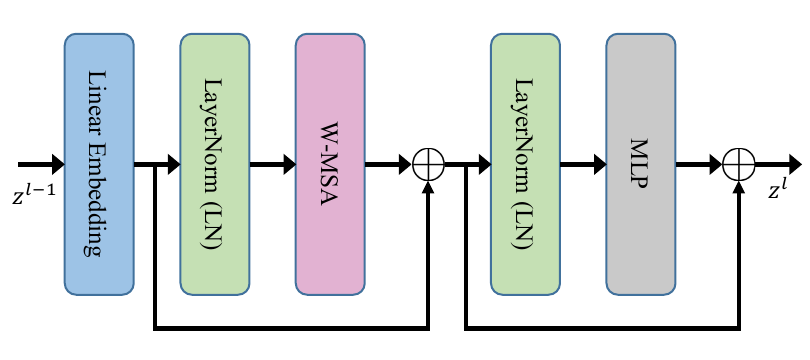}
    \caption{The structure of a Swin block. W-MSA is the Window-Multi head self-attention. MLP is the multilayer perceptron. }
    \label{fig123}
\end{figure}

Unlike previous denoising methods, image denoising via the noise-free style edit can be understood as an operation in the image style latent space, as shown in Fig. \ref{figlatent}. Previous methods focused on learning a better and more complex nonlinear mapping function in natural image manifolds. In contrast, we decouple the noise-free styles by connecting the image space and the style latent space through a style extractor and inject noise-free styles into noisy images' features for image denoising. In the following section, we give a detailed introduction to each component in our SDIDNet.

\subsection{Encoder-Decoder}
The encoder-decoder is a classic and powerful feature representation learner that has been widely used in image denoising tasks. In this work, we utilize the Transformer layer (i.e., Swin block \cite{l27}) to construct both the encoder and decoder.

\textbf{Swin Transformer:} Recently, Transformer has achieved great success in many computer vision tasks, mainly due to its powerful capacity in modeling long-range dependencies. However, self-attention in Transformer leads to a heavy computational burden. To alleviate this problem, Swin-Transformer with shifted and local window mechanism is proposed \cite{l27}.

The architecture of Swin block is shown in Fig. \ref{fig123}. Given the input feature $z^{l-1} \in R^{p\times q\times d}$, a linear embedding layer first reshapes the input feature into a feature of size $\frac{pq}{M^2} \times M^2 \times d$ through partitioning the input into non-overlapping local $M \times M$ windows, where $\frac{pq}{M^2}$ is the total number of partition windows. After that,  a LayerNorm (LN) layer is used to normalize all features of each sample, and the standard multi-head self-attention (MSA) is performed for each window. Specifically, for each local window feature $I \in R^{M^2 \times d}$ , the query, key and value matrices $Q,K,V \in R^{M^2 \times d}$ are calculated through learnable projection matrices $P_Q,P_K,P_V \in R ^{d \times d}$:
\begin{equation}
    Q = IP_Q, K = IP_K, V = IP_V
\end{equation}
where $P_Q,P_K,P_V$ are shared across all local windows. The attention matrix is then calculated for each window:
\begin{equation}
    MSA(Q,K,V) = softmax(\frac{QK^T}{\sqrt{d}}+bias)V,
\end{equation}
where $bias$ is the learnable relative positional encoding. After that, another LN followed by a multi-layer perceptron (MLP) is used for further feature transformation, where the MLP is composed of two fully connected (FC) layers and a GELU nonlinear activation. Skip connections are also applied for better information propagation. The pipeline of Swin block can be expressed as follows:
\begin{equation}
\begin{split}
         z^{l-1}  = & MSA(LN(z^{l-1})) + z^{l-1} \\
     z^l  = & MLP(LN(z^{l-1})) + z^{l-1}
\end{split}
\end{equation}

To further build connections among windows, the shifted window partition strategy was proposed in \cite{l27}, which is alternately used together with regular window partition.

\begin{figure*}[!t]
\centerline{\includegraphics[width=\textwidth]{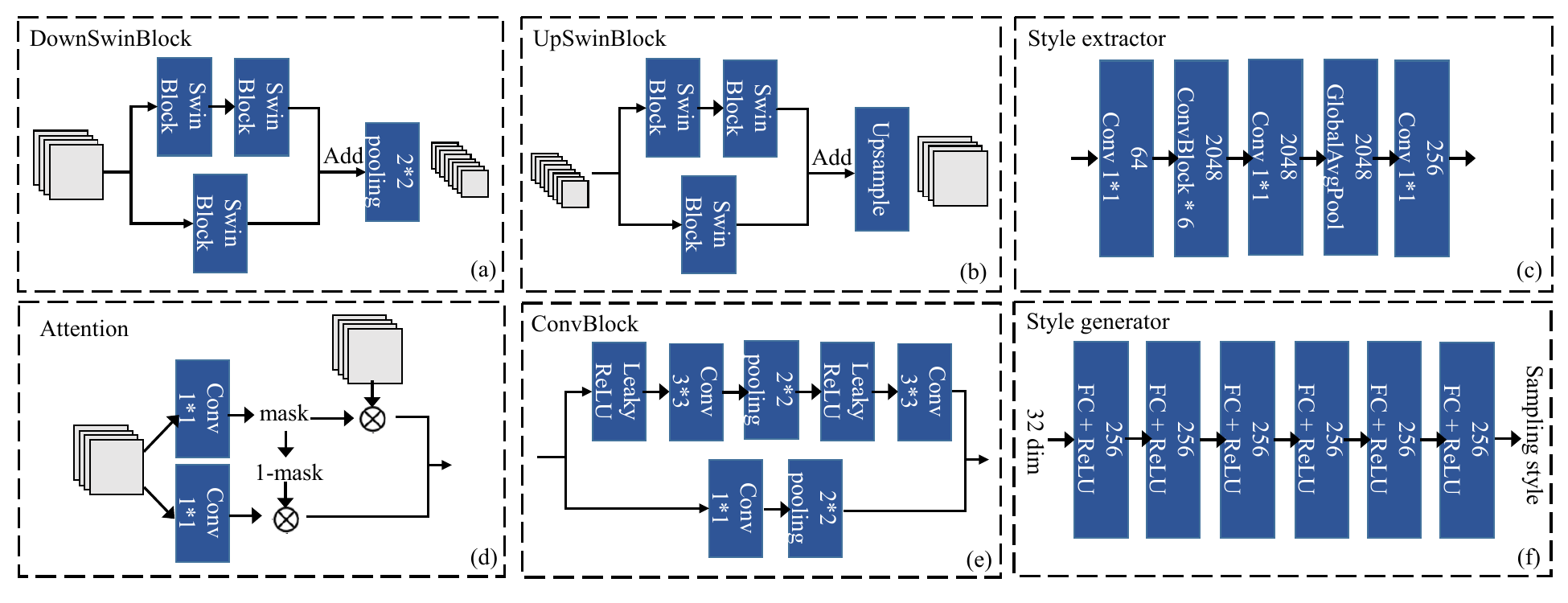}}
\caption{The detailed structure of the network modules. (a): DownSwinBlock module, (b): UpSwinBlock module, (c): Style extractor module, (d): Attention module, (e): ConvBlock module, (f): Style generator module.}
\label{fig2}
\end{figure*}

\textbf{Encoder:} It has been found that the convolution layer is effective in early visual processing, while Transformers are better suited for deep features \cite{l28}. Given a noisy image input $x \in \mathbb{R}^{H \times W \times 3} $ (H, W, and 3 are the image height, width, and channel number, respectively), we first use a $3 \times 3$ convolution layer to extract shallow features $F_{sf} \in \mathbb{R}^{H \times W \times C}$ from $x$, where $C$ is the channel number of shallow features. The $F_{sf}$ is then sent to the several cascaded DownSwinBlocks for deep features extraction. As shown in Fig. \ref{fig2} (a), a single DownSwinBlock module mainly comprises the two-branch Swin blocks. The outputs of the two branches are then element-wise added, followed by a $2 \times 2$ pooling layer to get low-resolution features. The output of DownSwinBlocks $f_{DSB}$ can be expressed as
\begin{equation}
    F_e = f_{DSB}(F_{sf})
\end{equation}

\textbf{Decoder:} The decoder is composed of several cascaded UPSwinBlocks  and a convolutional layer at the last. The structure of a single UPSwinBlock is similar to that of DownSwinBlock, as shown in Fig. \ref{fig2} (b). An upsampling layer is applied at the end of DownSwinBlock to double the feature resolution. The output of UPSwinBlocks $f_{USB}$ can be expressed as
\begin{equation}
    F_d = f_{USB}(F_{in})
\end{equation}
where $F_{in}$ is the input feature of the decoder, and $F_d$ is the output feature of the last UPSwinBlock module. At last, a $3 \times 3$ convolution operation on $F_d$ generates the final output $ F_{out}$ which has the same size as the noisy image. 

Note that during the training our encoder-decoder is only used to learn good feature representations of the inputs (both $x$ and $y$) rather than the mapping from $x$ to $y$ as done in current deep learning-based methods.

\subsection{Style extractor and Style generator}
The style extractor is developed to extract the noise and noise-free styles from noisy and clean images, respectively. As shown in Fig. \ref{fig2} (c), an input image is converted into deep features through multiple convolution layers, and then transformed into a 2048-dimension vector by global average pooling. Next, a $1 \times 1$ convolution is adapted to generate a 256-dimension vector, which represents the input image's style.

In the training phase, clean images are accessible to get the noise-free styles by feeding them to the style extractor while the noise-free styles are inaccessible in the inference phase due to the absence of clean images. To address this issue, a style generator is developed to learn the noise-free style distribution. The architecture of the style generator is shown in Fig. \ref{fig2}(f). Following styleGAN's style generation \cite{l11}, our style generator takes a 32-dimensional vector sampled from normal distribution as input and then passes it through six FC layers and the ReLU activation to generate a 256-dimensional style vector.

\subsection{Style conversion}
Style conversion (SC) is devised to remove the noise of intermediate features. It injects noise-free styles into the noisy images' features through a series of AdaINBlocks, each of which consists of an AdaIN operation and the convolutional layer. As shown in Fig. \ref{fig1}, the SC module starts with a $1 \times 1$ convolution layer that reduces the number of channels in the input feature. Next, the given style is passed through a fully connected layer to obtain adaptive parameters $\{s_s,s_b\}$, which are used to modulate the mean of the feature channels. Then, $N$ successive AdaIN operations \cite{l11} are employed to edit the feature. It has been demonstrated that feature statistics, such as mean and variance, carry the style information of an image \cite{l29}. Injecting a noise-free style will produce higher activations for image content features and lower activations for noise features, which means that the noise information in the feature channel is suppressed while the spatial structure of the content image is preserved.

To prevent the AdaIN operations from modifying the global attributes such as the content and background, we further design an attention module to preserve the image content attribute. The attention module operates on both channels and spatial directions, and the detailed structure is shown in Fig. \ref{fig2} (d). It can be defined as
\begin{equation}
    F_{sc} = Mask \odot F_{e} + (1 - Mask) \odot F_{adain}
\end{equation}
where $F_e$ is the feature obtained by the encoder, $F_{adain}$ is the feature after style editing, and $Mask$ is a mask generated from $F_{adain}$ through a $1 \times 1$ convolution layer and sigmoid function. $\odot$ denotes element-wise multiplication.

\subsection{Training loss function}
\textbf{Reconstruction loss.} 
We first design the following reconstruction loss: 
\begin{equation} \label{rec_loss}
\begin{aligned}
    \mathcal{L}_{rec} & = \lambda_1 \mathbb{E}_{p(x,y)} \left( \Vert x - dec(enc(x)) \Vert_1 \right. \\ & \left.+   \Vert y - dec(enc(y)) \Vert_1  \right. \\ 
                      &  \left. +  \Vert x - dec(sc(enc(x_i),s_{noise})) \Vert_1 \right) \\
                      & + \lambda_2 \mathbb{E}_{p(x,y)} \left (  \Vert y - dec(sc(enc(x), s_{noise\_free})) \Vert_1  \right. \\
                      & \left. +   \Vert y - dec(sc(enc(x), s_{gen})) \Vert_1 \right )
\end{aligned}
\end{equation}
where $s_{noise} = ext(x)$ is the noise style, $s_{noise\_free} = ext(y)$ is the noise-free style, $s_{gen}$ is the sampling style. The balance coefficients $\lambda_1$ and $\lambda_2$ are also included. The expectation $\mathbb{E}$ is taken over the joint distribution $p(x, y)$ and is practically approximated by the summation on the collected dataset.

In particular, the first two terms in Eq. \eqref{rec_loss} encourage the encoder-decoder to learn a feature representation without removing noise. The third and fourth terms encourage the style extractor to generate accurate style vectors, achieving style disentanglement. The $sc$ operation adjusts noisy image features to those corresponding to the given style. The last item encourages the style generator to learn an accurate distribution of noise-free styles.

\textbf{Style regression loss.} We additionally design a style regression loss, similar to \cite{l24}:
\begin{equation}
    \begin{split}
        \mathcal{L}_{sty} & = \mathbb{E}_{p(x,y)} \left ( \Vert s_{gen} - ext(x_{trg}) \Vert_1 \right. \\
        & \left. + \Vert s_{gen} - s_{noise\_free} \Vert_1 \right)
    \end{split}
\end{equation}
where $x_{trg}= dec(sc(enc(x), s_{gen}))$.

The first term in the style regression loss encourage the extracted style of the reconstruction image from the sampled styles to be consistent with the sampled style itself, and the second term encourages the style generator to generate a sampling style similar to the noise-free style.

\textbf{Full loss.} Finally, our full loss function can be written as
\begin{equation}
    \mathcal{L}_{full} = \mathcal{L}_{rec} + \lambda_{sty}\mathcal{L}_{sty}
\end{equation}
where $\mathcal{\lambda}_{sty}$ is the hyperparameter.

\section{Experiments and Evaluation}
In the following section, we provide a detailed overview of our experimental settings, present the experimental results, and demonstrate the effectiveness of SDIDNet through ablation experiments. We evaluate the performance of SDIDNet over two datasets (synthetic Gaussian noise dataset and two real-world image denoising datasets (SIDD \cite{l19}, DND \cite{l20})). Peak signal-to-noise ratio (PSNR) and structural similarity (SSIM) are adopted to evaluate denoising performance. 

\subsection{Training settings}
\label{setting}
SDIDNet is an end-to-end training model and requires no pre-training. In the training phase, the batch size was set to 6, and all images were randomly cropped to $128 \times 128$.  Hyperparameters in the reconstruction loss were set as follows: $\lambda_1 = 0.1, \lambda_2 = 0.3$. Hyperparameters in the full loss were set as follows: $\lambda_{sty} = 0.1$. We used two Adam optimizers with the first-order equilibrium constants $\beta_1=0.9$, and the second-order equilibrium constants $\beta_2=0.999$. The learning rate of the style generator was set to 1e-6, and the other learning rates were set to 2e-4. We used the cosine annealing strategy, and the learning rate gradually decayed from 2e-4 to 1e-6. At the same time, we used rotation, flipping, and cropping strategies for data enhancement. All experiments were done on a single RTX 2080Ti GPU.

\begin{figure}[!t]
    \includegraphics[width=\columnwidth]{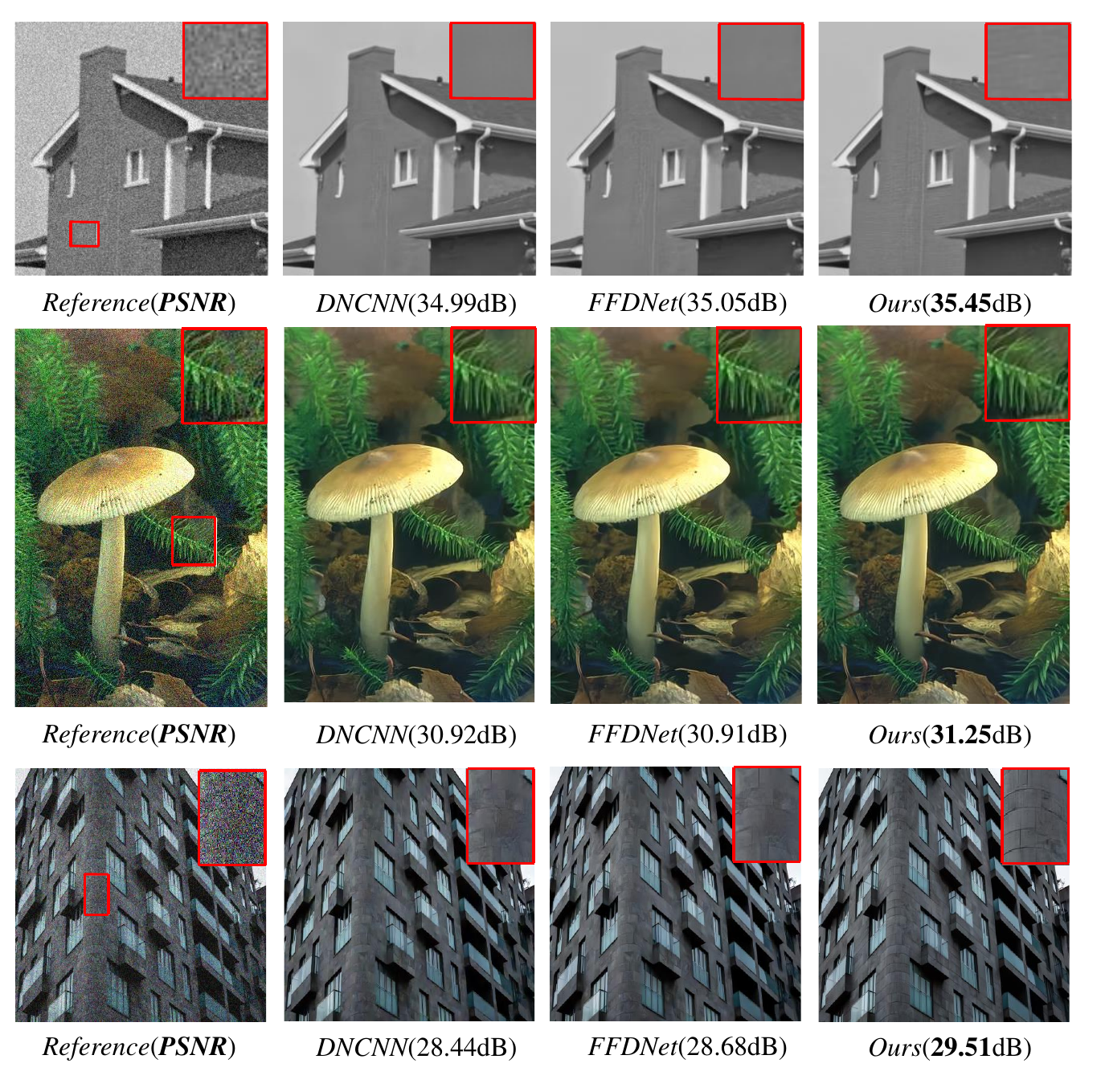}
    \caption{Visual results on synthetic Gaussian denoising. Top row: Gaussian grayscale denoising ($\sigma=15$). Middle row: Gaussian color denoising ($\sigma=25$). Bottom row: Gaussian color denoising ($\sigma=50$). The red box is the region of interest (ROI).}
    \label{figGaussian}
\end{figure}

\begin{table}[!t]
\begin{center}
\caption{Average PSNR(dB) of different methods with noise levels 15, 25 and 50 on the Set12 and BSD68.}
\label{gaussianGray}
\begin{tabular}{p{45pt}<{\centering} |p{20pt}<{\centering}  p{20pt}<{\centering}  p{20pt}<{\centering} |p{20pt}<{\centering}  p{20pt}<{\centering}  p{20pt}<{\centering}}
\hline
Dataset&& Set12& & & BSD68 & \\
\hline
$\sigma$ &15&25&50&15&25&50 \\
\hline
BM3D\cite{j37}& 32.37  & 29.97 & 26.72 & 31.08 & 28.57 & 25.60    \\
WNNM\cite{j38}& 32.70 & 30.28 & 27.05 & 31.37 & 28.83 & 25.87\\
DnCNN\cite{l1}& 32.86 & 30.44 & 27.18 & 31.73 & 29.23 & 26.23 \\
IRCNN\cite{j39}  & 32.77 & 30.38 & 27.14 & 31.63 & 29.15 & 26.19     \\
FFDNet\cite{l18} & 32.75 & 30.43 & 27.32 & 31.63 & 29.19 & 26.29    \\
Ours   & \textbf{33.01}  &   \textbf{30.69}& \textbf{27.61}  &   \textbf{31.80}& \textbf{29.35}  &   \textbf{26.45}   \\
\hline
\end{tabular}
\end{center}
\end{table}

\subsection{Results on synthetic Gaussian noise dataset}

We first evaluate SDIDNet on the synthetic Gaussian noise dataset. The training set consists of 900 images from the DIV2K dataset \cite{j30} and 2750 images from the Flick2K dataset\cite{j31} while the test set compries Set12\cite{l1}, BSD68\cite{j32}, CBSD68\cite{j33}, Kodak24\cite{j34}, McMaster\cite{j35}, and Urban100\cite{j36}. We follow the standard practice of adding additive Gaussian White noise (AWGN) with noise level $\sigma=15, 25, 50$ to both grayscale and color images. Additionally, we chose traditional denoising algorithms (BM3D\cite{j37}, WNNM\cite{j38}) and previous DNN-based denoisers (DnCNN\cite{l1}, IRCNN\cite{j39}, FFDNet\cite{l18}) as the comparison methods.

To evaluate the effectiveness of SDIDNet, we measure the PSNR and SSIM of all images on the test set and then average the results. Experimental results are shown in Table \ref{gaussianGray} and Table \ref{gaussianColor}. It can be seen that SDIDNet has achieved the best PSNR at different noise levels and on different test sets. Specifically, for the grayscale image test sets (Set12, BSD68), the PSNR of SDIDNet is about 0.3dB and 0.1dB higher than that of DnCNN at three noise levels, respectively. The first row in Fig. \ref{figGaussian} shows the denoising results of DnCNN, FFDNet, SDIDNet on the image "house" from Set12. It is apparent that SDIDNet may display clearer texture details. For the color image test sets (CBSD68, Kodak24, McMaster, Urban100), SDIDNet is more competitive than the previous method. The second and third rows in Fig. \ref{figGaussian} show the denoising results at noise levels of 25 and 50, respectively. It can be observed that SDIDNet has significant visual advantages.

\begin{table}[!t]
\begin{center}
\caption{Average PSNR(dB) of different methods with noise levels 15, 25 and 50 on the CBSD68, Kodak24, McMaster and Urban100.}
\label{gaussianColor}
\begin{tabular}{p{25pt}<{\centering} p{20pt}<{\centering} | p{20pt}<{\centering}  p{20pt}<{\centering} p{20pt}<{\centering}  p{20pt}<{\centering}  p{20pt}<{\centering}}
\hline
Dataset&$\sigma$ & CBM3D\cite{j37} & DnCNN\cite{l1}& IRCNN\cite{j39} & FFDNet \cite{l18}& Ours \\
\hline
&15&33.52&33.90&33.86&33.87& \textbf{34.14}\\
CBSD68&25&30.71&31.24&31.16&31.21&\textbf{31.51}\\
&50&27.38&27.95&27.86&27.96&\textbf{28.34}\\
\hline
&15&34.28&34.60&34.69&34.63&\textbf{34.96}\\
Kodak24&25&32.15&32.14&32.18&32.13&\textbf{32.55}\\
&50&28.46&28.95&28.93&28.98&\textbf{29.50}\\
\hline
&15&34.06&33.45&34.58&34.66&\textbf{35.00}\\
McMaster&25&31.66&31.52&32.18&32.35&\textbf{32.76}\\
&50&28.51&28.62&28.91&29.18&\textbf{29.75}\\
\hline
&15& 32.58 & 32.98 & 33.78 & 33.83 &\textbf{34.27}\\
Urban100&25 & 30.68& 30.81 & 31.20 & 31.40&\textbf{31.97}\\
&50&27.41&27.59&27.70&28.05&\textbf{28.86}\\

\hline
\end{tabular}
\end{center}
\end{table}

\begin{table*}[!t]
\begin{center}
\caption{PSNR, SSIM, and computational cost of different methods on the SIDD and DND testsets.  Multiply–accumulate operations (MACs) and the number of parameters (Params) are estimated under the input size of 3 × 256 × 256. “Infer time” presents the inference time of an image with shape $3 \times 256 \times 256$.}
\label{table2}
\begin{tabular}{p{55pt}<{\centering} p{35pt}<{\centering} p{35pt}<{\centering} | p{35pt}<{\centering} p{35pt}<{\centering} | p{35pt}<{\centering}  p{40pt}<{\centering} p{70pt}<{\centering} }
\hline
&SIDD& &DND & & Cost& \\
Method  & PSNR(dB) & SSIM & PSNR(dB) & SSIM & MACs (G) & Params (M) & Infer Time (ms)\\
\hline
DnCNN \cite{l1} & 23.66  & 0.583 & 32.43  & 0.790  & 43.9  & 0.7 & 87.7    \\
BM3D \cite{j32} &  25.65 &    0.641 & 34.51  & 0.851 & - & - & -  \\
CBDNet  \cite{l21}  & 30.78  &    0.801 & 38.06  & 0.942 & 40.3  & 4.4 & 80.8   \\
RIDNet  \cite{l6} & 38.71  & 0.951   & 39.26  & 0.953 & 88.2  & 1.5   & 176.5   \\
AINDNet \cite{l22}  & 38.95  &0.952   & 39.37  & 0.951  & -  & 13.8  & -   \\
VDN  \cite{l32} & 39.28 &     0.956 & 39.38  & 0.952 & 41.9  & 7.8 &  82.8  \\ 
MPRNet\cite{l2} & 39.71    & 0.958  & 39.80  & 0.954 & 573.5 &   15.7 & 1147.1 \\
Ours   & 39.45  &  0.958  & 39.67  & 0.952   & \textbf{28.9}  & 12.7  & \textbf{57.9} \\
\hline
\end{tabular}
\end{center}
\end{table*}

\subsection{Results on SIDD Benchmark}
\label{sidd}

The Smart Phone Image Denoising Data Set (SIDD) is a set of noisy images obtained in 10 scenes using five smartphones under different lighting conditions. We use 320 high-resolution color images for training, while SIDD provides 1280 color images($128 \times 128$) for the test set. We choose the previous DnCNN \cite{l1}, BM3D \cite{j32}, CBDNet \cite{l21}, RIDNet \cite{l6}, AINDNet \cite{l22}, VDN \cite{l32}, and MPRNet \cite{l2} methods for comparison. Table \ref{table2} lists the PSNR and SSIM results.  

\begin{figure}[!t]
\includegraphics[width=\columnwidth]{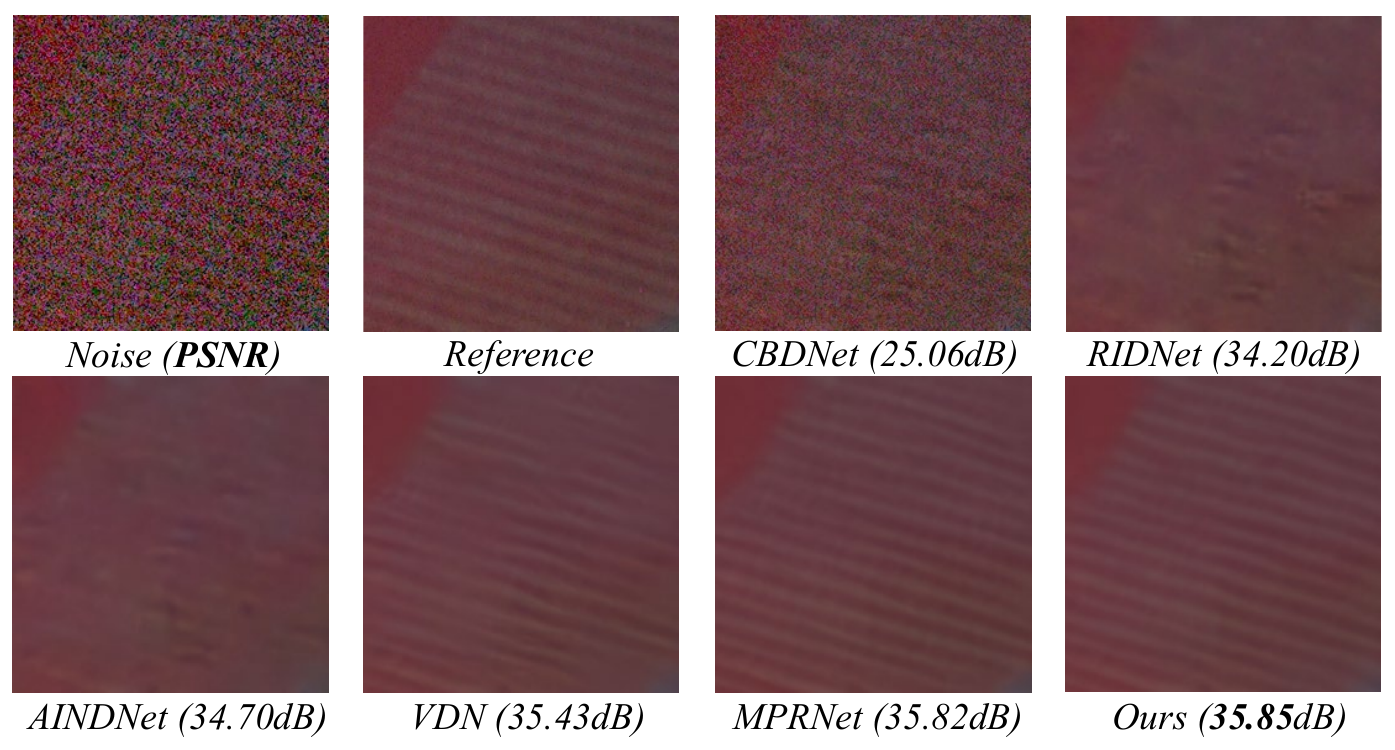}
\caption{Visual comparisons of different methods on SIDD Benchmark. }
\label{fig4}
\end{figure}

Fig. \ref{fig4} displays results for a visual inspection. In this dataset, SDIDNet achieves a PSNR of 39.45 dB, which is 0.17 dB higher than that of VDN. As shown in Fig. \ref{fig4}, SDIDNet can effectively remove noise and retain texture details. MPRNet had a PSNR value of 39.71 dB, slightly higher (0.26dB) than SDIDNet. It is worth noting that MPRNet employs a multi-stage learning strategy, which builds a multi-level architecture to preserve contextual features, while SDIDNet uses a single-stage learning strategy in its current version.

We compute the computational cost of different methods and select images of size $3\times256\times256$ for testing. Multiply–accumulate operations (MACs) and inference time are shown in Table \ref{table2}. Surprisingly, our model can complete inference using only \textbf{5.0\%} of the MACs of MPRNet, which is mainly due to two reasons. Firstly, we use a single-stage training strategy rather than a two-stage one. Secondly, the depth of our Swin block is only 2 or 4, which is shallower than the standard Swin-Tiny \cite{l27}.

\begin{figure*}[!t]
    \includegraphics[width=\textwidth]{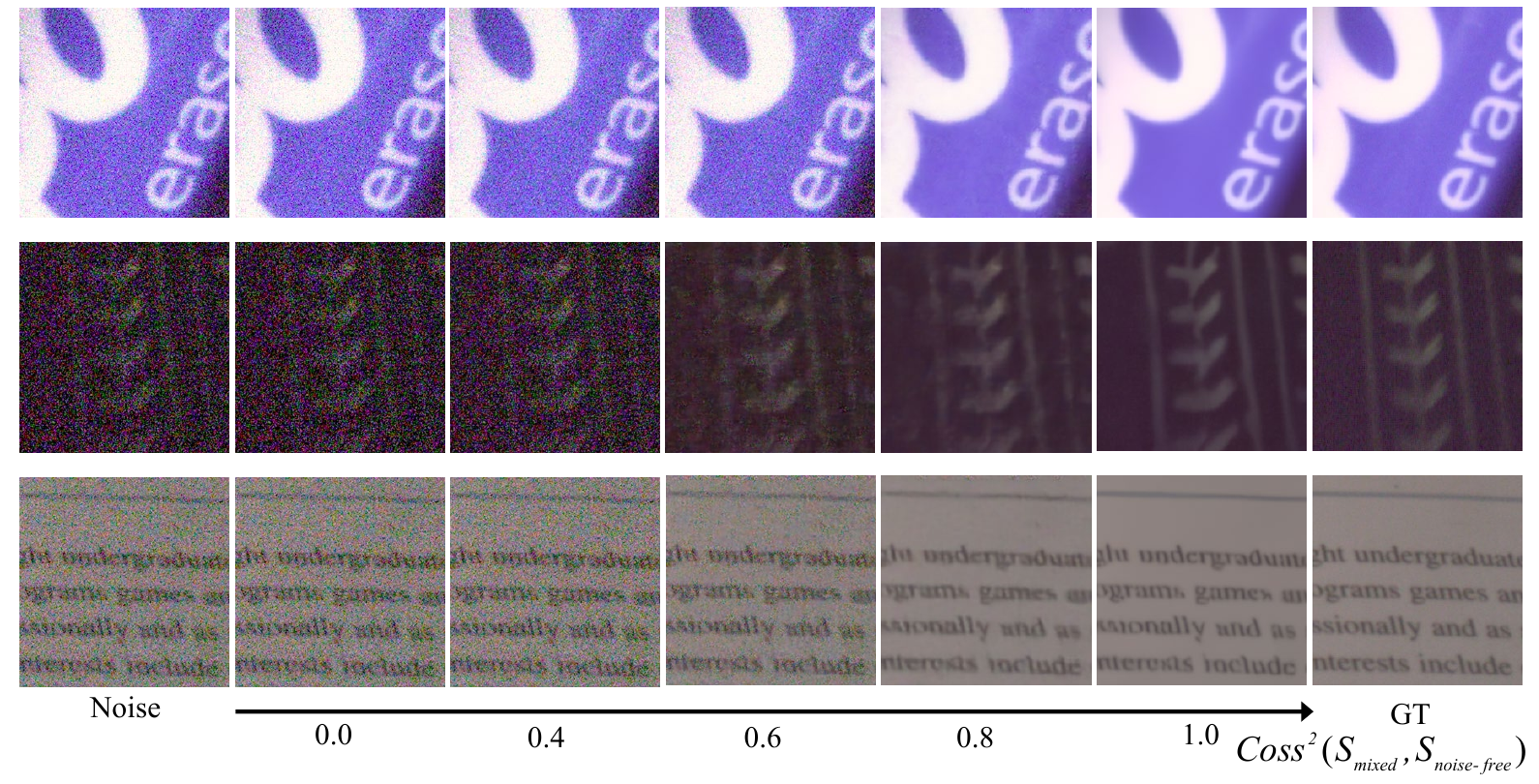}
    \caption{Visualization of denoising from mixed styles on SIDD testset. The first and last columns are noisy and clean images, and the middle five columns are denoised images with increasing $Coss^2(s_{mixed},s_{noise\_free})$, where $Coss$ denotes the cosine similarity measure.}
    \label{fig7}
\end{figure*}

\textbf{Mixing styles.} 
We present a detailed explanation of the denoising process involved in mixed-style editing, as shown in Fig. \ref{fig7}. Since the parameters of the style extractor are shared, both the noise style and the noise-free style are in the same style latent space. Define the mixed style as an intermediate state $s_{mixed} = \lambda s_{noise\_free} + (1-\lambda) s_{noise}$. By adjusting the parameter $\lambda$, we present the transition between the noise and noise-free styles, which corresponds to different degrees of denoising results.
It is evident that when only the noisy style is used ($\lambda=0$), the denoising result is almost identical to the input noisy image. As $\lambda$ gradually increases, the mixed style is changing from a noise style to a noise-free style, and the denoised image becomes progressively clearer. Such experiment demonstrates the good interpretability of our proposed method. 

\begin{figure}
    \centering
    \includegraphics[width=0.75\columnwidth]{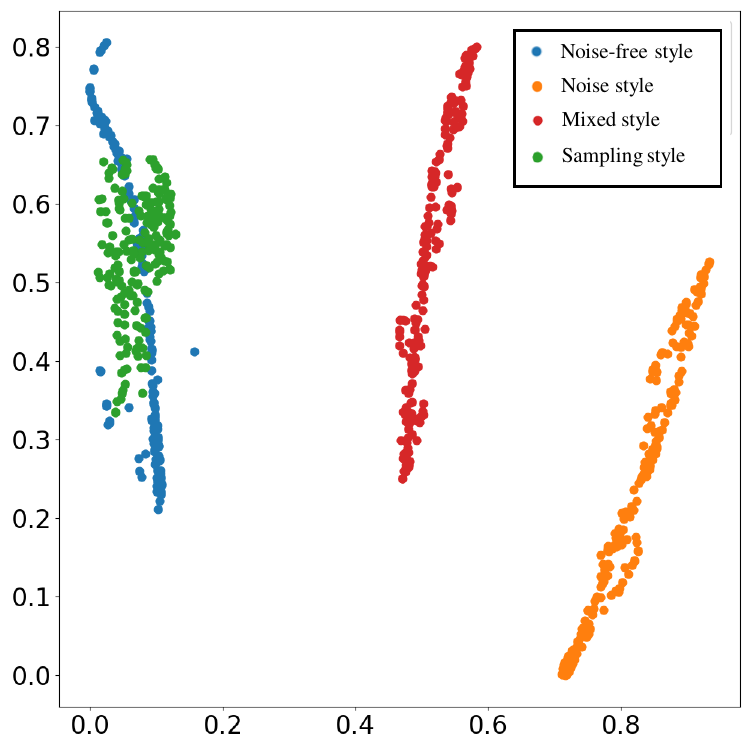}
    \caption{2D distribution map of noise-free style, noise style, sampling style, and mixed style using t-SNE algorithm.}
    \label{figstyledis}
\end{figure}

\textbf{Style distribution.} It is expected that the noise and noise-free styles should be easily distinguished in the style latent space, and the distribution of sampling styles should closely match that of noise-free styles. To demonstrate this, we perform a visualization analysis on these styles. Specifically, the styles are chosen as 256-dim vectors, and we use the t-SNE \cite{l39} algorithm to reduce their dimensionality into a two-dimensional embedding space for visualization purposes. The visualization results on the SIDD test set are shown in Fig. \ref{figstyledis}. The results demonstrate the following:

1)	The distributions of both noise-free and noise styles exhibit significant differences, indicating that the style extractor maps the image manifold to an accurate style latent space.

2)	The distribution of the sampling style is very close to that of the noise-free style, illustrating the effectiveness of our reconstruction loss and style regression loss.

3)	The mixed style is in between the noise-free and noise styles. Therefore, the mixed style can be regarded as building a pipeline from the noise styles to noise-free styles. The closer it is to the noise-free style, the lower the activation response to noise features, enabling controllable image denoising.

\subsection{Results on DND Benchmark}
\label{dnd}

\begin{figure}[!t]
\includegraphics[width=\columnwidth]{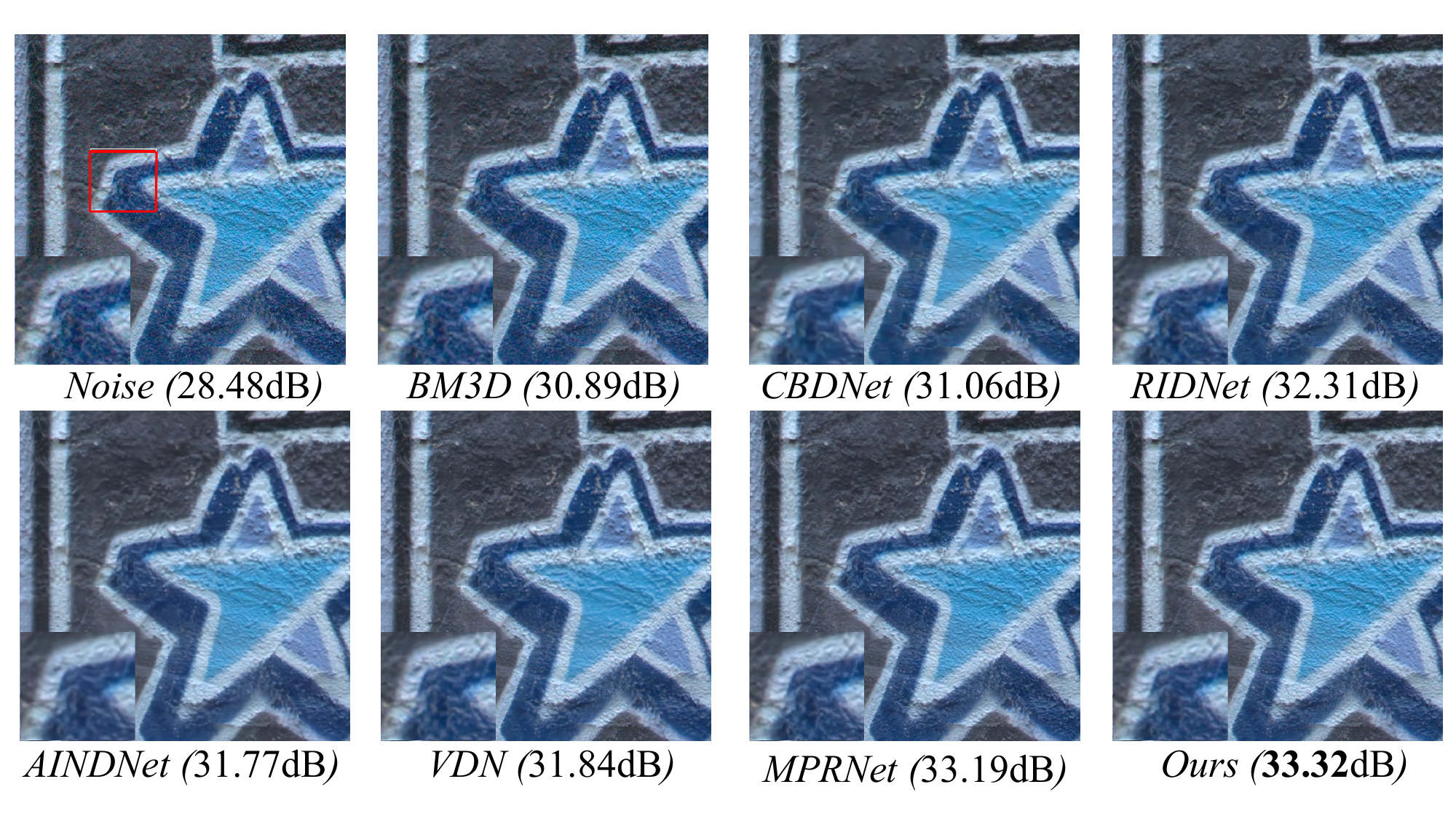}
\caption{Visual comparisons of different methods on DND Benchmark. The red box indicates the ROI. }
\label{fig5}
\end{figure}

The Darmstadt Noise Data Set (DND) comprises 50 pairs of real noisy images. For each pair, a reference image is captured with the base ISO level, while the noisy image is captured with a higher ISO and appropriately adjusted exposure time. Note that the DND dataset does not contain any training data. Therefore, we use the model trained on the SIDD train set and directly test on the DND benchmark, which can validate the performance of our method on different datasets.

The comparison results are shown in Table \ref{table2}, and we also provide a visual comparison in Fig. \ref{fig5}. PSNR of SDIDNet is 0.29dB higher than that of VDN, reflecting the effectiveness of our method.

\subsection{Ablation study}

\begin{figure}[!t]
    \centering
    \includegraphics[width=0.9\columnwidth]{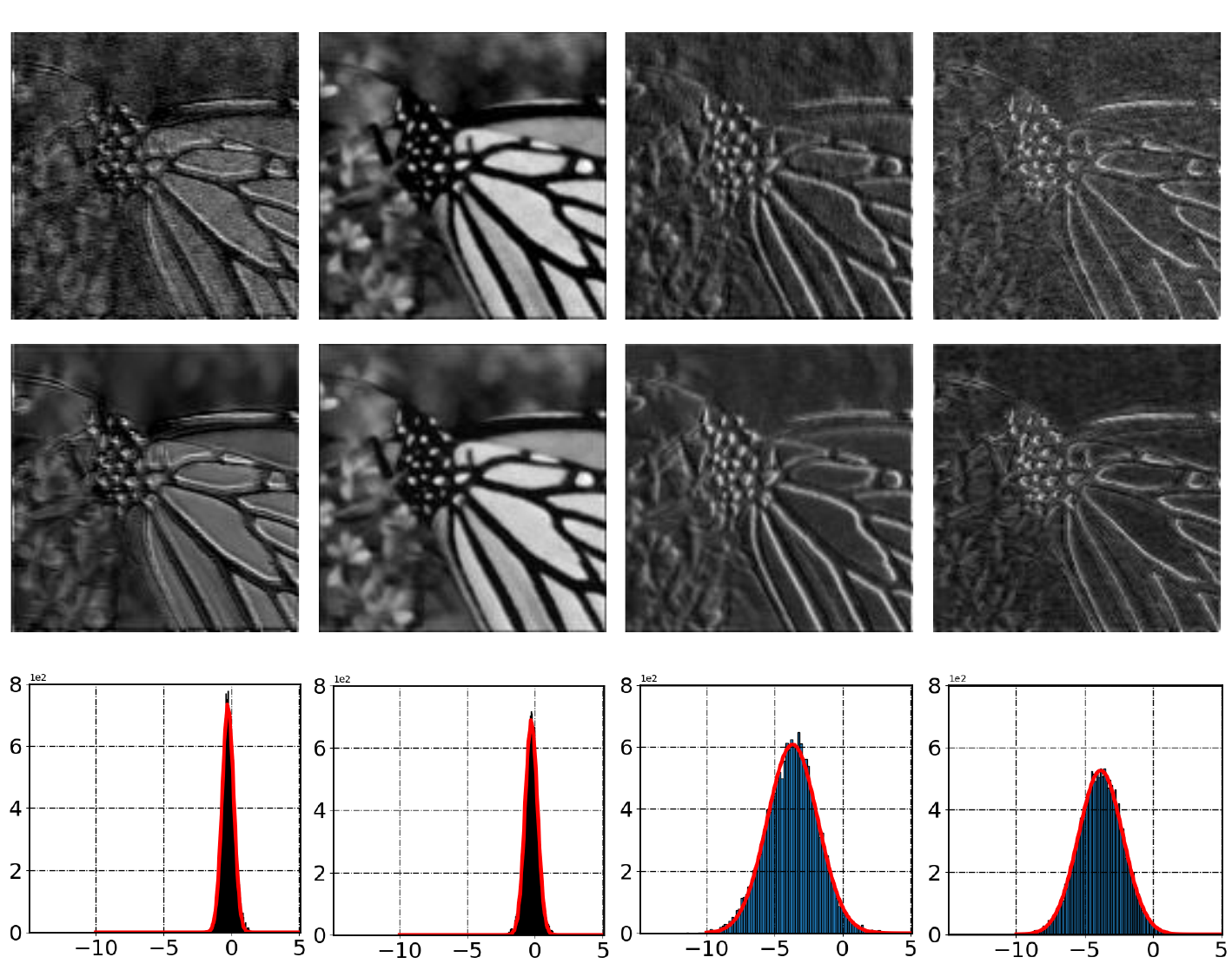}
    \caption{Image features and their differences before and after applying sampling style editing. Top row: features of the noisy image, middle row: features after sampling style editing. Bottom row: the feature difference between the first two rows. The blue and black line is the histogram of the feature difference, and the red line displays the fitted Gaussian curve after the difference. Each column of features belongs to the same channel.}
    \label{fig6}
\end{figure}

\begin{table}[!t]
\centering
\caption{PSNR(dB) with or without SC module on SIDD and DND.}
\begin{tabular}{p{60pt}<{\centering} p{30pt}<{\centering} p{40pt}<{\centering}}
\hline
Dataset  & with\_SC & without\_SC \\
\hline
SIDD  & 39.55   &   22.48  \\
DND    & 39.67  &   30.21   \\
\hline
\end{tabular}
\label{tablesc}
\end{table}

\textbf{Ablation on style conversion module.} Without retraining, we test the PSNR of denoised images with or without style conversion(SC) on the SIDD, and DND. Results are listed in Table \ref{tablesc}. We notice that, without the SC module, the SDIDNet degrades to a standard encoder-decoder, and its performance drops significantly. Besides, without the SC module, the denoised image from SDIDNet is very similar to the input noisy image, which indicates that the SDIDNet without the SC module actually learns the identity map. 

Simultaneously, we visualize the image features before and after applying sampling style editing, as shown in Fig. \ref{fig6}. The first row presents the features of the noisy "Monarch" image ($\sigma = 25$) from Set12. The second row displays the features after applying sampling style editing, and the third row shows the histogram and the corresponding fitted Gaussian curve of the feature difference. Features in the same column come from the same channel. The first row shows that the encoder implicitly encodes the image content features (the first two columns) and noisy features (the last two columns), respectively. The image features in the second row are significantly clean, indicating that the sampling style editing process can effectively remove noise. The difference histogram in the third row can be well-fitted to a Gaussian distribution, implying that the sampling style mainly removes the Gaussian-like signal of the features. Furthermore, for the first two columns of content feature channels, the mean of the difference histogram is close to 0, and the mean of the feature remains almost unchanged, indicating that the sampling style will generate a high-responsive activation of content features and preserve image features. For the last two columns of noisy feature channels, the feature differences are more dispersed, and the mean value of the histogram is significantly less than 0, indicating that the sampling style produces low-response activations to noise features and removes noise.

\begin{table}[!t]
\centering
\caption{Effects of the number of AdainBlock modules N on SIDD.}
\begin{tabular}{p{35pt}<{\centering} p{20pt}<{\centering} p{20pt}<{\centering}p{20pt}<{\centering} p{20pt}<{\centering}p{20pt}<{\centering}}
\hline
N & N=1 &  N=4 & N=8 & N=16 & N=32\\
\hline
PSNR(dB) &39.10 & 39.24  &\textbf{39.45} & 39.27 & 39.08\\
\hline
\end{tabular}
\label{table99}
\end{table}

\textbf{Influence of the AdainBlock number in the SC module.} Table \ref{table99} provides results obtained on SIDD with different numbers of AdainBlock. With N=8 as the inflection point, the denoising performance increases and then decreases. On the one hand, when N=1, the style set $\{s_s,s_b\}$ generated by the 256-dimensional style is only 64 groups, and it is challenging to decouple the content and the noise features. On the other hand, when N = 16 or 32, the higher dimensionality of the style set may increase the difficulty of model matching. In addition, we also test SDIDNet's denoising performance when keeping only $s_s$ and $s_b$ separately. Results show that with only the bias parameter $s_b$, SDIDNet fails to converge, and with only the scale parameter $s_s$, the performance of SDIDNet slightly decreases. For example, on SIDD, the PSNR drops to 39.14dB, which suggests that the process of editing the feature means is the primary factor for denoising, and $s_b$ can further improve the denoising ability.

\section{Discussion and Conclusion}
In this paper, we propose a novel strategy for image denoising based on style disentanglement and provide a new prospect in the image denoising task. Rather than relying on complex network architecture and accurate image noise modeling, the proposed style conversion operation can naturally convert noisy images' noise style into the noise-free style. We demonstrate the effectiveness of the style disentanglement strategy in image denoising tasks. In addition, we experimentally find that the noise-free style produces high response activations to content features and low response activations to noise features, thereby removing noise. We believe that image denoising via style disentanglement provides an interpretable framework and its application to other low-level image reconstruction tasks is worthy of further exploration.

\begin{figure}
    \centering
    \includegraphics[width=\columnwidth]{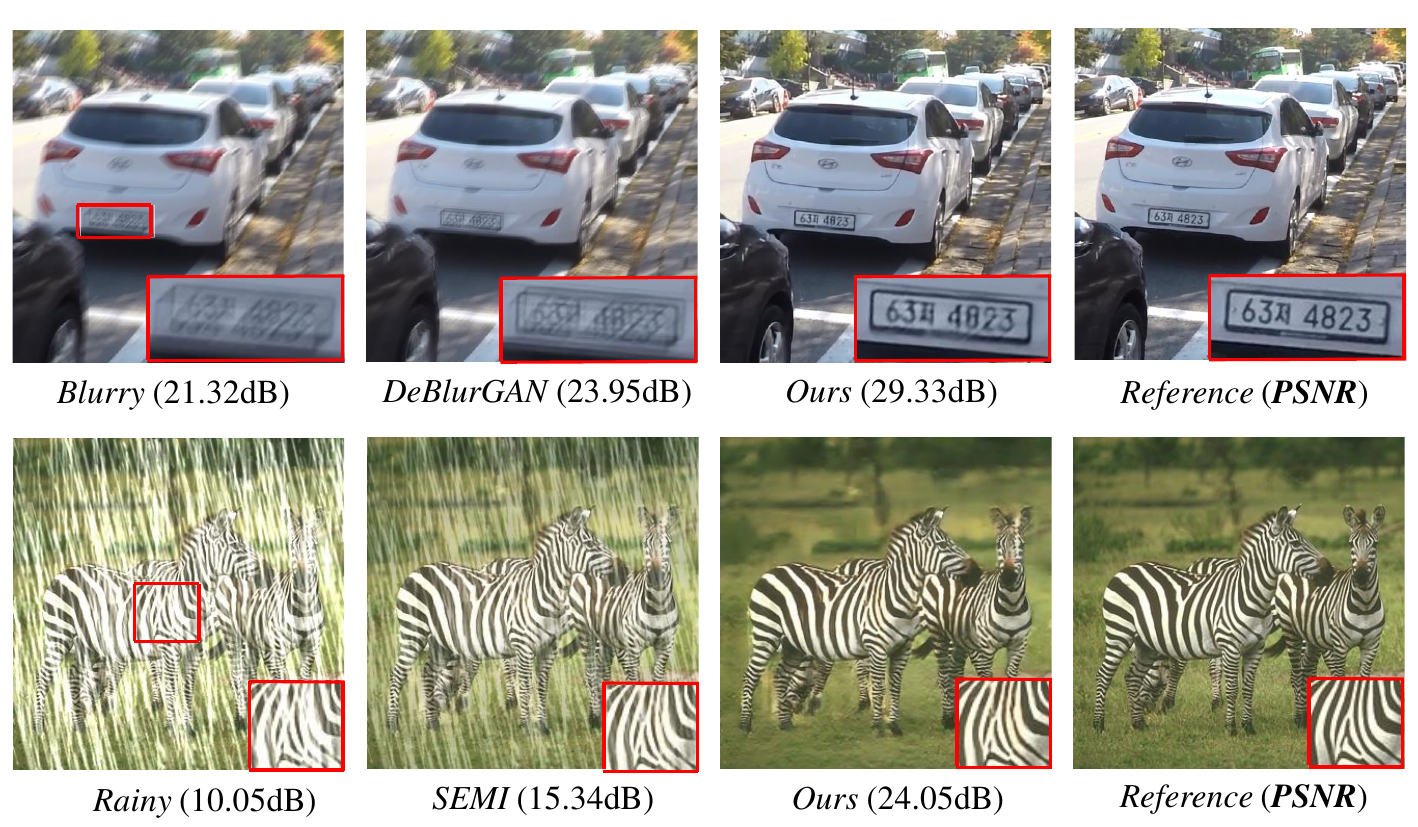}
    \caption{Comparison of different methods on the image motion deblur (top row) and image derain task (bottom row). The red boxes are the ROI.}
    \label{derain}
\end{figure}

\begin{table}[!t]
\begin{center}
\caption{Average PSNR(dB) of different methods on the image motition deblur and image derain.}
\label{derainTable}
\begin{tabular}{p{30pt}<{\centering} | p{30pt}<{\centering} | p{25pt}<{\centering}   p{30pt}<{\centering}  p{25pt}<{\centering}  p{25pt}<{\centering}}
\hline
Task&Dataset& U10SR\cite{l40} & DeblurGAN\cite{l41} & SVR \cite{l42}& Ours \\
\hline
Deblur&GoPro&21.99&28.70&29.19&\textbf{29.53}\\
\hline
& & DerainNet \cite{l43}& SEMI \cite{l44}& UMRL\cite{l45} &  \\
\hline
&Rain100H&14.92&16.56&26.01& \textbf{26.85}\\
Derain&Rain100L&27.03&25.03&29.18&\textbf{33.32}\\
&Test1200&23.38&26.05&\textbf{30.55}&30.46\\
\hline
\end{tabular}
\end{center}
\end{table}

Here, we show some initial experimental results of SDIDNet on other tasks instead of image denoising. We apply the SDIDNet to image motion deblurring and image deraining. The U10SR \cite{l40}, DeblurGAN \cite{l41}, SVR \cite{l42}, DerainNet \cite{l43}, SEMI \cite{l44}, and UMRL \cite{l45} methods as employed for comparison. The results are shown in Fig. \ref{derain} and Table \ref{derainTable}.

For the image deblurring task, blurred and clean images have distinct unique blur and sharp features, corresponding to blur and sharp styles, respectively. We evaluate the SDIDNet on the GoPro dataset, and the average PSNR is 29.53 dB, 0.83 dB higher than that of the classic DeblurGAN \cite{l41}. We also observe that, while editing blurred image features with the sharp style, the sharp style will produce low-response activation to blurred features, achieving deblurring.

Similarly, for the image deraining task, rainy and clean images have unique rainy and clean image features, corresponding to rainy and rain-free styles, respectively. The rain-free style produces low-response activation for rainy features, and SDIDNet outperforms the classic DerainNet \cite{l43} and SEMI \cite{l44} algorithms on the Rain100H, Rain100L, and Test1200 datasets. These results demonstrate that the style disentanglement strategy widely applies to various low-level vision tasks.

We note that our current SDIDNet model requires a supervised training strategy, which may restrict its application in real-world scenarios where large paired data is difficult to collect. Further research on how to extract noise-free styles from noisy images is a promising direction. In addition, our paper did not consider and employ the latest Transformer blocks introduced in \cite{Restormer, EEM}, leading to a performance gap compared with state-of-the-art methods. Incorporating these modules into our encoder-decoder is the future research.

\end{document}